\documentclass{article}

\usepackage[final]{neurips_2023}
\usepackage{natbib}
\usepackage[utf8]{inputenc} 
\usepackage[T1]{fontenc}    
\usepackage{hyperref}       
\usepackage{url}            
\usepackage{booktabs}       
\usepackage{amsfonts}       
\usepackage{nicefrac}       
\usepackage{microtype}      
\usepackage{xcolor}         
\usepackage{bbm}
\usepackage{amsmath}
\usepackage{bm}
\usepackage{graphicx}
\usepackage{caption}
\usepackage{subcaption}

\title{Hierarchical Vision Transformers for Context-Aware Prostate Cancer Grading in Whole Slide Images}

\author{%
	Clément Grisi\thanks{Corresponding author: \texttt{clement.grisi@radboudumc.nl}} , Geert Litjens, Jeroen van der Laak \\
	\\
	Computational Pathology Group, Radboudumc, Netherlands \\
}

\begin{document}

	\maketitle

	\begin{abstract}	
		Vision Transformers (ViTs) have ushered in a new era in computer vision, showcasing unparalleled performance in many challenging tasks. However, their practical deployment in computational pathology has largely been constrained by the sheer size of whole slide images (WSIs), which result in lengthy input sequences. Transformers faced a similar limitation when applied to long documents, and Hierarchical Transformers were introduced to circumvent it. Given the analogous challenge with WSIs and their inherent hierarchical structure, Hierarchical Vision Transformers (H-ViTs) emerge as a promising solution in computational pathology. This work delves into the capabilities of H-ViTs, evaluating their efficiency for prostate cancer grading in WSIs. Our results show that they achieve competitive performance against existing state-of-the-art solutions.
	\end{abstract}

	\section{Introduction}
	
	With the advent of whole-slide imaging technology, large numbers of tissue slides are being scanned and archived digitally. Digitization of histological slides has contributed to the growing availability of large datasets, fostering tremendous computer vision research opportunities. Recent progress in deep learning has set new performance standards in various clinical applications, especially with the recent surge of attention-based models (\cite{vit}).\\
	\\
	However, conventional deep learning methods are ill-equipped to handle the enormous sizes of whole-slide images (WSIs), which do not fit into the memory of graph processing units (GPUs). Innovative strategies have emerged to overcome this memory bottleneck. A prevailing approach consists of partitioning these massive images into smaller patches. These patches often serve as the input units, with labels derived from pixel-level annotations. Because the traditional requirement of input-label pairs can be prohibitive – or even impossible as pathologists can only annotate what they know – recent research has explored techniques beyond full supervision, focusing on more flexible training paradigms, such as weakly supervised learning. Multiple Instance Learning (MIL) has recently emerged as a powerful weakly supervised approach in several computational pathology challenges, where it has shown remarkable performance (\cite{Hou2015EfficientMI,campanella2019}).\\
	\\
	Despite showcasing impressive results, most MIL methods neglect the spatial relationships among patches, thereby overlooking valuable contextual information. To address this limitation, recent research focused on developing methods capable of integrating context (\cite{sparseconvmil,streaming,transmil}). By leveraging the hierarchical structure inherent to WSIs, Hierarchical Vision Transformers (H-ViTs) have proven successful at learning context-aware image representations from whole slide images, achieving state-of-the-art results in cancer subtyping and survival prediction (\cite{hipt}). We explore their application in the multi-class classification setting of prostate cancer grading. We additionally provide enhanced model interpretability by introducing an innovative approach for computing factorized attention heatmaps. Our code is available at \href{https://github.com/computationalpathologygroup/hvit}{\texttt{https://github.com/computationalpathologygroup/hvit}}.

	\section{Proposed Method}
	\label{sec:methods}
	
	\paragraph{Hierarchical Vision Transformer.} The inherent hierarchical structure within whole-slide images spans across various scales, from tiny cell-centric patches containing fine-grained information, all the way up to the entire slide, which exhibits overall intra-tumoral heterogeneity of the tissue microenvironment. Along this spectrum, various patch sizes may capture cell-to-cell interactions or macro-scale interactions. Drawing inspiration from this layered structure, our H-ViT model mimicks of the multi-stage architecture introduced in \cite{hipt}, which incorporates multiple Vision Transformers. Each performs bottom-up aggregation, effectively mapping the token sequence at one scale into a single representation at the subsequent scale, eventually resulting in a slide-level feature vector (Appendix \ref{appendix:hvit}). A first Vision Transformer (referred to as patch-level Transformer) performs cell-level aggregation by breaking down $256\times256$ patches into $16\times16$ mini-patches. Then, a second Transformer (referred to as region-level Transformer) builds context-aware embeddings by aggregating non-overlapping $256\times256$ patches within larger $2048\times2048$ regions. Finally, a third Transformer (referred to as slide-level Transformer) pools the resulting region-level tokens into a single slide-level representation that can be used for downstream prediction tasks.\\
	\\
	We experimented with two variants of this model. The first, deemed Global H-ViT, replicates HIPT (\cite{hipt}) : both patch-level and region-level Transformers are pretrained and kept frozen. Only the slide-level Transformer is finetuned on the downstream task. The second one, deemed Local H-ViT, consists of a pretrained and frozen patch-level Transformer, with both the region-level and slide-level Transformers finetuned on the downstream task.
	
	\paragraph{Refined Attention Factorization.} It's important to underscore that the attention scores from the different Transformers hold varying degrees of relevance in highlighting the areas critical for the model's predictions. Indeed, some are pretrained and frozen, while others are finetuned for the downstream classification task. To address this disparity, we introduce a parameter $\gamma \in \left[0, 1\right]$ to balance the influence of attention scores between frozen and finetuned Transformers.\\
	\\
	Let $N$ be the total number of Transformers involved in the pretraining or the finetuning processes. Let $n$ be the number of frozen Transformers. We denote by $a^i_{(x,y)}$ the attention score of the $i$-th Transformer $\text{T}_i$ for the pixel with $(x,y)$ coordinates in the slide. The factorized attention score for that pixel, $a_{(x,y)}$, is then computed as:
	
	\begin{equation} \label{eq:blend_attention}
		a_{(x,y)} = \frac{1}{\beta} \sum_{i=0}^{N-1} a^i_{(x,y)} \big( \gamma \mathbbm{1}_{\text{F}}(\text{T}_i) + (1-\gamma) (1-\mathbbm{1}_{\text{F}}(\text{T}_i)) \big)
	\end{equation}
	\\
	\\
	\vspace{-1cm}\\
	\\
	with $\beta = n\cdot(1-\gamma) + (N-n)\cdot \gamma$ and $ \mathbbm{1}_{\text{F}}(\text{T}_i) $ the indicator function defined on the set of Transformers $\left\{\text{T}_i\right\}_{i=0}^{N-1}$ and equal to $1$ if $\text{T}_i$ is finetuned, $0$ otherwise. This flexible approach allows for fine-grained control over how attention scores are combined, making it possible to emphasize either pretrained or task-specific features in the model's attention mechanism.

	\section{Experimental Results}
	\label{sec:results}
	
	\paragraph{Dataset.} To assess the robustness of the proposed method, we use the PANDA dataset, introduced in \cite{panda}. It is the largest publicly available dataset of H\&E stained prostate WSIs to date, with $11,554$ prostate biopsies curated from two different sites. All slides are provided at a pixel spacing close to $0.50$ $\mu$m, with their corresponding ISUP score (Appendix \ref{appendix:panda}).
	
	\paragraph{Data Preprocessing \& Evaluation Metric.} We adapted \cite{clam} preprocessing pipeline to automatically segment tissue in each slide, from which we extract non-overlapping $2048\times2048$ regions at a resolution of $0.50$ micron per pixel (Appendix \ref{appendix:hs2p}). We split the development set into $5$ cross-validation folds, stratifying on the ISUP score. To evaluate the model’s classification performance, we report quadratic weighted kappa scores on the tuning set, the public test set and the private test set, averaged across the $5$ folds.
	
	\paragraph{Prostate Cancer Grading.} We pretrain the patch-level and region-level Transformers on the PANDA dataset via the student-teacher knowledge distillation framework DINO (\cite{dino}). To leverage the ordinal nature of the ISUP scores, we formulated the classification problem as a regression task and used the Mean Squared Error (MSE) loss. Classification results are summarized in Table \ref{tab:results}. Local H-ViT achieves higher macro-averaged performance than Global H-ViT and comparable performance with the winning solutions to the PANDA challenge (Appendix \ref{appendix:winning_panda}). Allowing gradients to flow through the region-level Transformer grants the model more freedom to refine the pretrained features, such that they better fit the classification task at hand.

	\begin{table}[h]
		\caption{Classification results}
		\label{tab:results}
		\centering
		\begin{tabular}{lcccc}
			\\ [-1.mm] 
			\toprule
			Model & \# parameters & Tune Score &  Public Test Score &  Private Test Score \\
			\midrule
			Global H-ViT  & $594,818$ & $0.888$ ± $0.019$  & $0.799$ ± $0.018$ & $0.807$ ± $0.016$ \\
			Local H-ViT & $3,348,098$ & $\bm{0.950}$ \textbf{±} $\bm{0.002}$  & $\bm{0.904}$ \textbf{±} $\bm{0.011}$ & $\bm{0.904}$ \textbf{±} $\bm{0.011}$ \\
			\bottomrule
		\end{tabular}
	\end{table}
	
\begin{figure}[h]
		\centering
		\begin{subfigure}[h]{0.23\textwidth}
			\centering
			\includegraphics[height=1.6\textwidth]{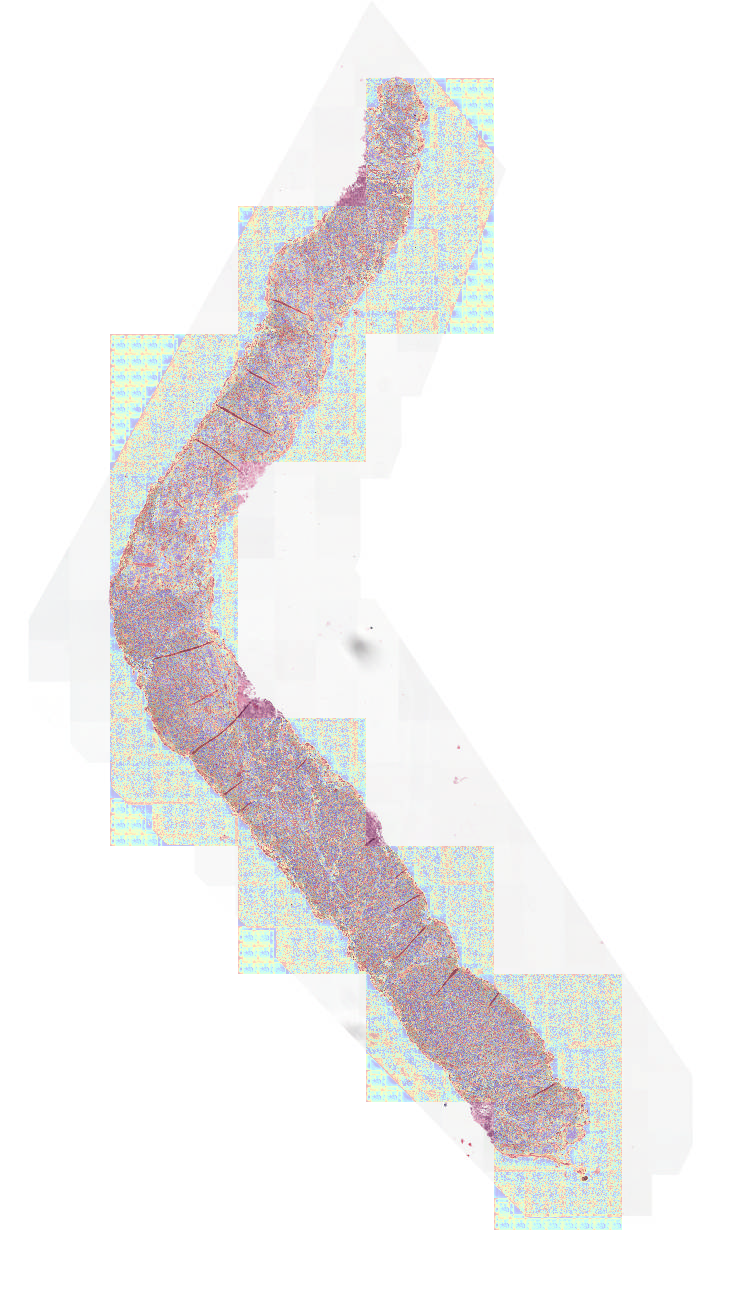}
			\caption{patch-level}
		\end{subfigure}
		\begin{subfigure}[h]{0.23\textwidth}
			\centering
			\includegraphics[height=1.6\textwidth]{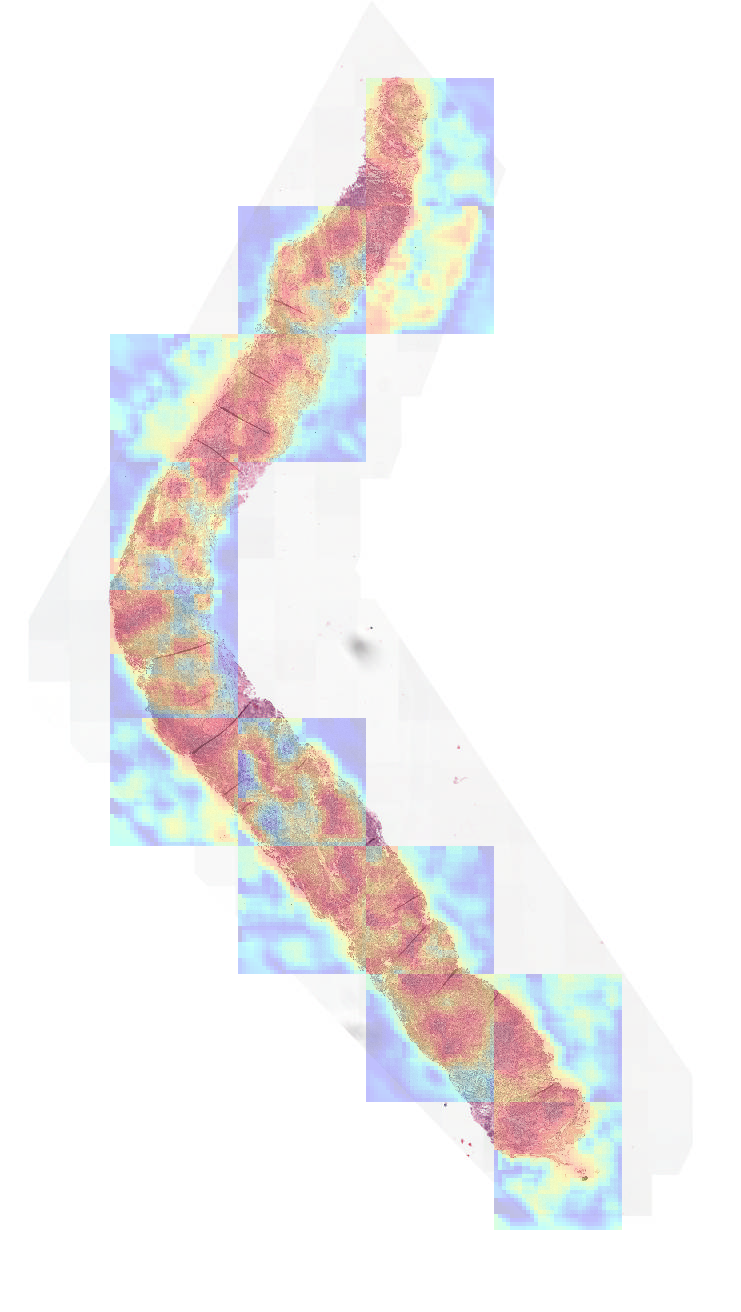}
			\caption{region-level}
		\end{subfigure}
		\begin{subfigure}[h]{0.23\textwidth}
			\centering
			\includegraphics[height=1.6\textwidth]{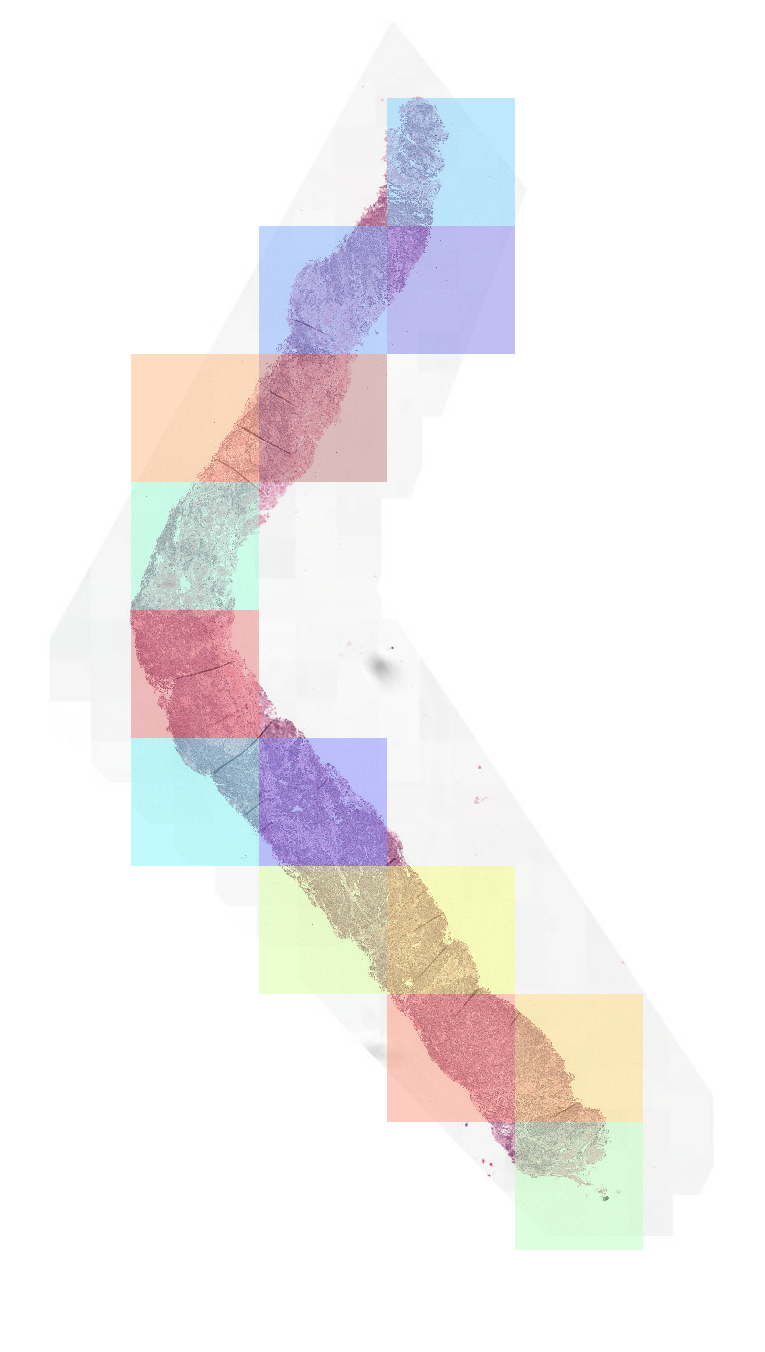}
			\caption{slide-level}
		\end{subfigure}
		\begin{subfigure}[h]{0.23\textwidth}
			\centering
			\includegraphics[height=1.6\textwidth]{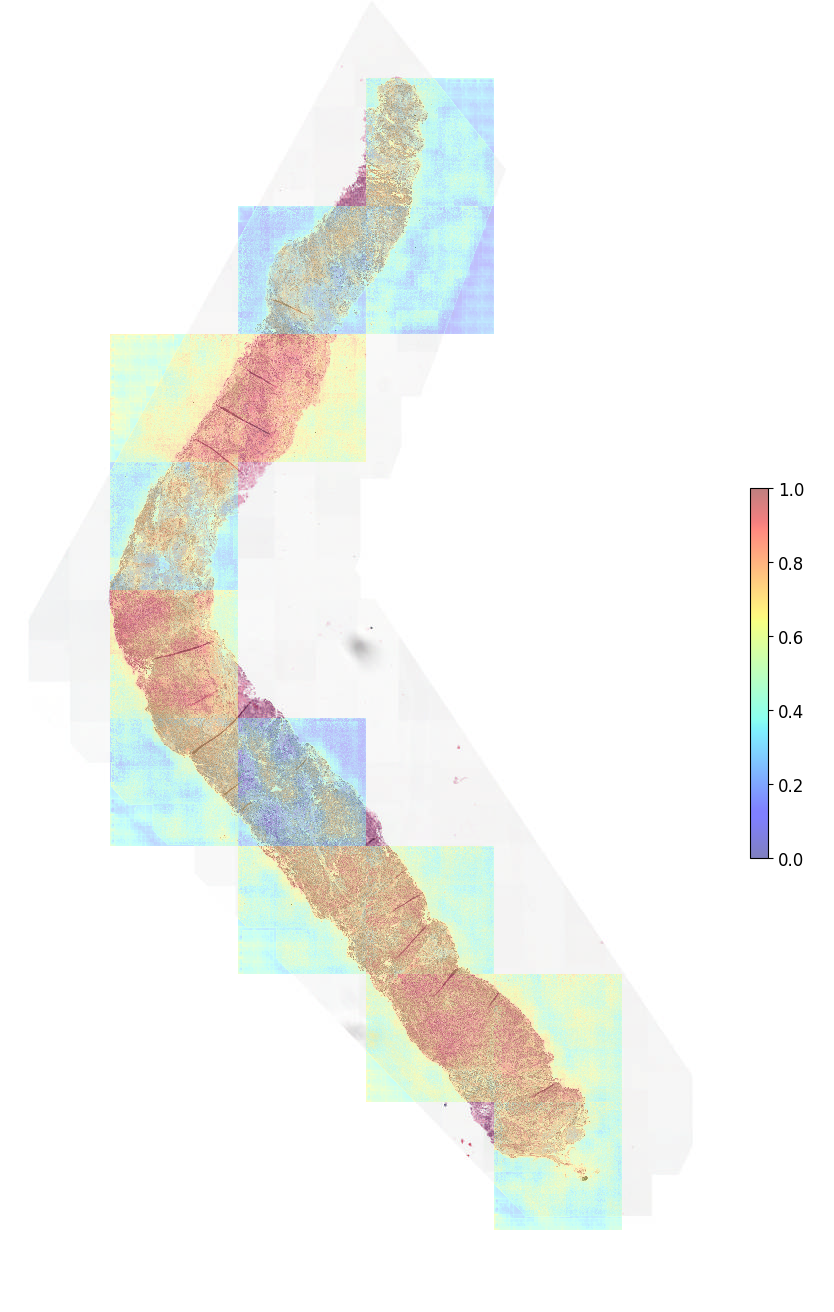}
			\caption{factorized ($\gamma = 0.7$)}
		\end{subfigure}
		\caption{Local H-ViT individual and factorized attention heatmaps}
		\label{fig:hm}
	\end{figure}

	\paragraph{Model Interpretability.} Attention heatmaps offer a streamlined form of model interpretability, revealing the specific image features that the model has learned to associate with each class. To that end, factorized heatmaps generated using our approach offer a more comprehensive view of model attention than heatmaps from single-level Transformers (Figure \ref{fig:hm}). Using $\gamma = 0.5$ in Equation \ref{eq:blend_attention} equally weights the contributions of the frozen and finetuned Transformers. We advocate for $\gamma$ values greater than $0.5$, as these upweight the contributions of finetuned attention scores, hence highlighting features that are more directly related to the prediction task at hand. Among the various values tested, $\gamma = 0.7$ yielded visually satisfying and informative heatmaps, balancing fine-grained and coarse-grained attention features (Appendix \ref{appendix:hm}). We only show heatmaps for Local H-ViT ($n=1$) as it is our best-performing model.
	
	

	\section{Conclusion}
	
	We presented Hierarchical Vision Transformer (H-ViT), a promising method to address the unique challenges associated with the analysis of whole-slide images in computational pathology. By leveraging the inherent hierarchical structure of WSIs, H-ViT efficiently captures context-aware representations, providing a comprehensive view of the tissue microenvironment. Our experiments showcased the efficacy of this method in prostate cancer grading, with Local H-ViT outperforming Global H-ViT and achieving performance comparable to the winning solutions of the PANDA challenge. Our proposed attention factorization method offers a nuanced and controllable mechanism for model interpretability, providing practical insights that could be valuable for clinical interpretation.  
	
%
	\section*{Acknowledgments}
	
	 We express our sincere gratitude to Richard Chen for his pioneering work, "Scaling Vision Transformers to Gigapixel Images via Hierarchical Self-Supervised Learning". The model adopted in this paper is inspired by his research. We are especially thankful for his generous support and timely answers, which were instrumental in enhancing our understanding and successful implementation of the model.
	
	\section*{Potential Negative Societal Impact}
	
	To the best of our knowledge, we did not find any potential negative societal impact of our research. All clinical data used in our experiments were sourced from anonymized, publicly available datasets, ensuring patient privacy. Our method could help doctors make faster and more consistent decisions, in the interest of patients. It is not meant to conduct any direct diagnosis.
	
	{\small
		\bibliography{egbib}

\begin{thebibliography}{10}
\providecommand{\natexlab}[1]{#1}
\providecommand{\url}[1]{\texttt{#1}}
\expandafter\ifx\csname urlstyle\endcsname\relax
  \providecommand{\doi}[1]{doi: #1}\else
  \providecommand{\doi}{doi: \begingroup \urlstyle{rm}\Url}\fi

\bibitem[Bulten et~al.(2022)Bulten, Kartasalo, Chen, Str{\"o}m, Pinckaers,
  Nagpal, Cai, Steiner, van Boven, Vink, Hulsbergen-van~de Kaa, van~der Laak,
  Amin, Evans, van~der Kwast, Allan, Humphrey, Gr{\"o}nberg, Samaratunga,
  Delahunt, Tsuzuki, H{\"a}kkinen, Egevad, Demkin, Dane, Tan, Valkonen,
  Corrado, Peng, Mermel, Ruusuvuori, Litjens, Eklund, Brilhante,
  {\c{C}}ak{\i}r, Farr{\'e}, Geronatsiou, Molini{\'e}, Pereira, Roy, Saile,
  Salles, Schaafsma, Tschui, Billoch-Lima, Pereira, Zhou, He, Song, Sun,
  Yoshihara, Yamaguchi, Ono, Shen, Ji, Roussel, Zhou, Chai, Weng, Grechka,
  Shugaev, Kiminya, Kovalev, Voynov, Malyshev, Lapo, Campos, Ota, Yamaoka,
  Fujimoto, Yoshioka, Juvonen, Tukiainen, Karlsson, Guo, Hsieh, Zubarev,
  Bukhar, Li, Li, Speier, Arnold, Kim, Bae, Kim, Lee, Park, and
  consortium]{panda}
Wouter Bulten, Kimmo Kartasalo, Po-Hsuan~Cameron Chen, Peter Str{\"o}m, Hans
  Pinckaers, Kunal Nagpal, Yuannan Cai, David~F. Steiner, Hester van Boven,
  Robert Vink, Christina Hulsbergen-van~de Kaa, Jeroen van~der Laak, Mahul~B.
  Amin, Andrew~J. Evans, Theodorus van~der Kwast, Robert Allan, Peter~A.
  Humphrey, Henrik Gr{\"o}nberg, Hemamali Samaratunga, Brett Delahunt, Toyonori
  Tsuzuki, Tomi H{\"a}kkinen, Lars Egevad, Maggie Demkin, Sohier Dane, Fraser
  Tan, Masi Valkonen, Greg~S. Corrado, Lily Peng, Craig~H. Mermel, Pekka
  Ruusuvuori, Geert Litjens, Martin Eklund, Am{\'e}rico Brilhante, Asl{\i}
  {\c{C}}ak{\i}r, Xavier Farr{\'e}, Katerina Geronatsiou, Vincent Molini{\'e},
  Guilherme Pereira, Paromita Roy, G{\"u}nter Saile, Paulo G.~O. Salles, Ewout
  Schaafsma, Jo{\"e}lle Tschui, Jorge Billoch-Lima, Em{\'i}io~M. Pereira, Ming
  Zhou, Shujun He, Sejun Song, Qing Sun, Hiroshi Yoshihara, Taiki Yamaguchi,
  Kosaku Ono, Tao Shen, Jianyi Ji, Arnaud Roussel, Kairong Zhou, Tianrui Chai,
  Nina Weng, Dmitry Grechka, Maxim~V. Shugaev, Raphael Kiminya, Vassili
  Kovalev, Dmitry Voynov, Valery Malyshev, Elizabeth Lapo, Manuel Campos,
  Noriaki Ota, Shinsuke Yamaoka, Yusuke Fujimoto, Kentaro Yoshioka, Joni
  Juvonen, Mikko Tukiainen, Antti Karlsson, Rui Guo, Chia-Lun Hsieh, Igor
  Zubarev, Habib S.~T. Bukhar, Wenyuan Li, Jiayun Li, William Speier, Corey
  Arnold, Kyungdoc Kim, Byeonguk Bae, Yeong~Won Kim, Hong-Seok Lee, Jeonghyuk
  Park, and the PANDA~challenge consortium.
\newblock Artificial intelligence for diagnosis and gleason grading of prostate
  cancer: the panda challenge.
\newblock \emph{Nature Medicine}, 28\penalty0 (1):\penalty0 154--163, Jan 2022.
\newblock ISSN 1546-170X.
\newblock \doi{10.1038/s41591-021-01620-2}.
\newblock URL \url{https://doi.org/10.1038/s41591-021-01620-2}.

\bibitem[Campanella et~al.(2019)Campanella, Hanna, Geneslaw, Miraflor, Werneck
  Krauss~Silva, Busam, Brogi, Reuter, Klimstra, and Fuchs]{campanella2019}
Gabriele Campanella, Matthew~G. Hanna, Luke Geneslaw, Allen Miraflor, Vitor
  Werneck Krauss~Silva, Klaus~J. Busam, Edi Brogi, Victor~E. Reuter, David~S.
  Klimstra, and Thomas~J. Fuchs.
\newblock Clinical-grade computational pathology using weakly supervised deep
  learning on whole slide images.
\newblock \emph{Nature Medicine}, 25\penalty0 (8):\penalty0 1301--1309, Aug
  2019.
\newblock ISSN 1546-170X.
\newblock \doi{10.1038/s41591-019-0508-1}.
\newblock URL \url{https://doi.org/10.1038/s41591-019-0508-1}.

\bibitem[Caron et~al.(2021)Caron, Touvron, Misra, J\'egou, Mairal, Bojanowski,
  and Joulin]{dino}
Mathilde Caron, Hugo Touvron, Ishan Misra, Herv\'e J\'egou, Julien Mairal,
  Piotr Bojanowski, and Armand Joulin.
\newblock Emerging properties in self-supervised vision transformers.
\newblock In \emph{Proceedings of the International Conference on Computer
  Vision (ICCV)}, 2021.

\bibitem[Chen et~al.(2022)Chen, Chen, Li, Chen, Trister, Krishnan, and
  Mahmood]{hipt}
Richard~J. Chen, Chengkuan Chen, Yicong Li, Tiffany~Y. Chen, Andrew~D. Trister,
  Rahul~G. Krishnan, and Faisal Mahmood.
\newblock Scaling vision transformers to gigapixel images via hierarchical
  self-supervised learning.
\newblock In \emph{Proceedings of the IEEE/CVF Conference on Computer Vision
  and Pattern Recognition (CVPR)}, pages 16144--16155, June 2022.

\bibitem[Dosovitskiy et~al.(2021)Dosovitskiy, Beyer, Kolesnikov, Weissenborn,
  Zhai, Unterthiner, Dehghani, Minderer, Heigold, Gelly, Uszkoreit, and
  Houlsby]{vit}
Alexey Dosovitskiy, Lucas Beyer, Alexander Kolesnikov, Dirk Weissenborn,
  Xiaohua Zhai, Thomas Unterthiner, Mostafa Dehghani, Matthias Minderer, Georg
  Heigold, Sylvain Gelly, Jakob Uszkoreit, and Neil Houlsby.
\newblock An image is worth 16x16 words: Transformers for image recognition at
  scale, 2021.

\bibitem[Hou et~al.(2015)Hou, Samaras, Kurç, Gao, Davis, and
  Saltz]{Hou2015EfficientMI}
Le~Hou, Dimitris Samaras, Tahsin~M. Kurç, Yi~Gao, James~E. Davis, and
  J.~Saltz.
\newblock Efficient multiple instance convolutional neural networks for
  gigapixel resolution image classification.
\newblock \emph{ArXiv}, abs/1504.07947, 2015.
\newblock URL \url{https://api.semanticscholar.org/CorpusID:16405142}.

\bibitem[Lerousseau et~al.(2021)Lerousseau, Vakalopoulou, Deutsch, and
  Paragios]{sparseconvmil}
Marvin Lerousseau, Maria Vakalopoulou, Eric Deutsch, and Nikos Paragios.
\newblock Sparseconvmil: Sparse convolutional context-aware multiple instance
  learning for whole slide image classification, 2021.

\bibitem[Lu et~al.(2021)Lu, Williamson, Chen, Chen, Barbieri, and
  Mahmood]{clam}
Ming~Y. Lu, Drew F.~K. Williamson, Tiffany~Y. Chen, Richard~J. Chen, Matteo
  Barbieri, and Faisal Mahmood.
\newblock Data-efficient and weakly supervised computational pathology on
  whole-slide images.
\newblock \emph{Nature Biomedical Engineering}, 5\penalty0 (6):\penalty0
  555--570, Jun 2021.
\newblock ISSN 2157-846X.
\newblock \doi{10.1038/s41551-020-00682-w}.
\newblock URL \url{https://doi.org/10.1038/s41551-020-00682-w}.

\bibitem[Pinckaers et~al.(2021)Pinckaers, Bulten, Van~der Laak, and
  Litjens]{streaming}
Hans Pinckaers, Wouter Bulten, Jeroen Van~der Laak, and Geert Litjens.
\newblock Detection of prostate cancer in whole-slide images through end-to-end
  training with image-level labels.
\newblock \emph{IEEE Trans Med Imaging}, PP, March 2021.
\newblock ISSN 1558-254X.
\newblock \doi{10.1109/TMI.2021.3066295}.

\bibitem[Shao et~al.(2021)Shao, Bian, Chen, Wang, Zhang, Ji, and
  Zhang]{transmil}
Zhuchen Shao, Hao Bian, Yang Chen, Yifeng Wang, Jian Zhang, Xiangyang Ji, and
  Yongbing Zhang.
\newblock Transmil: Transformer based correlated multiple instance learning for
  whole slide image classification, 2021.

\end{thebibliography}
	}
	
	\clearpage
	\appendix
	
	\section{Architecture Overview}
	\label{appendix:hvit}
	
\begin{figure}[h]
		\centering
		\includegraphics[width=0.97\textwidth]{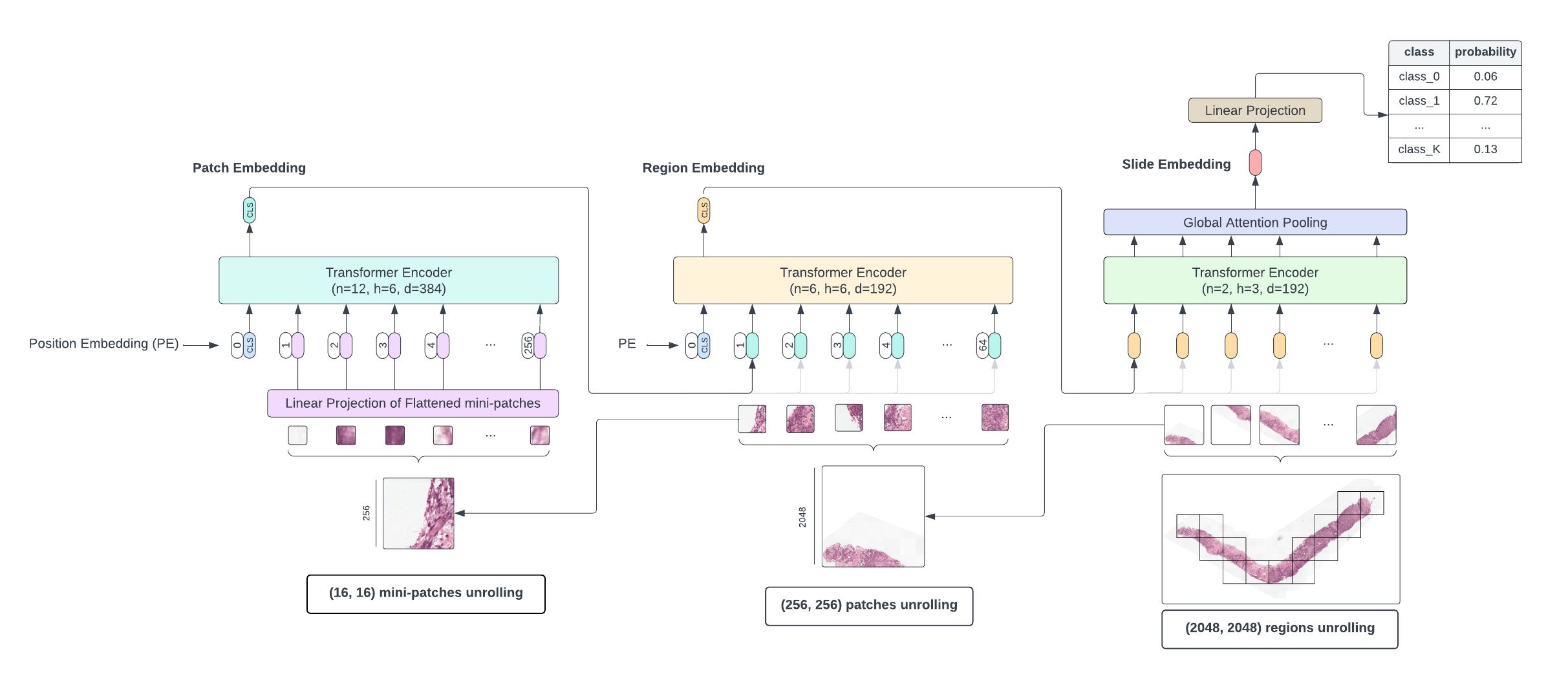}
		\caption{H-ViT overview}
		\label{fig:hvit}
	\end{figure}

	Figure \ref{fig:hvit} shows the multi-stage H-ViT architecture we use in this work. It features three Vision Transformers, followed by a simple linear classifier that projects the resulting slide embedding onto the desired number of classes.
	
	\vspace{-4mm}
	
	\section{PANDA Dataset Details}
	\label{appendix:panda}
	
	In Table \ref{tab:panda}, we provide a summary of the main characteristics of the PANDA dataset.
	
	\vspace{-1mm}
	
	\begin{table}[h]
		\caption{PANDA dataset summary}
		\label{tab:panda}
		\centering
		\begin{tabular}{llcccc}
			\\ [-1.1mm] 
			\toprule
			Site& Scanner & Spacing ($\mu$m) & \# dev  & \# public test  & \# private test \\
			\midrule
			Radboud & 3DHistech  & $0.48$ & $5160$ & $195$ & $333$ \\
			Karolinska     & Leica & $0.50$ & $2193$ & $97$ & $150$ \\
			Karolinska     & Hamamatsu  & $0.45$ & $3263$ & $101$ & $62$ \\
			\bottomrule
		\end{tabular}
	\end{table}

\begin{figure}[!h]
		\centering
		\includegraphics[width=0.65\textwidth]{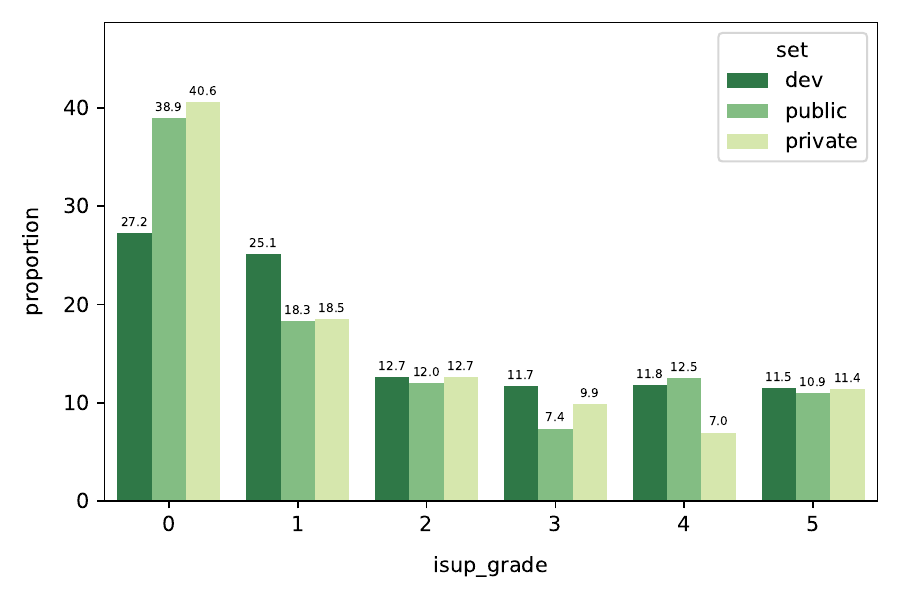}
		\caption{PANDA label distribution}
		\label{fig:label_distribution}
	\end{figure}

	Pathologists classify tumors into different growth patterns by analyzing the histological architecture of the tumor tissue. Tissue specimens are then categorized into one of five groups based on the distribution of these patterns in the tumor. Figure \ref{fig:label_distribution} shows the grade group distribution for the development set, the public test set and the private test set.
	
	\section{Data Preprocessing}
	\label{appendix:hs2p}
	
\begin{figure}[!h]
		\centering
		\begin{subfigure}{0.45\textwidth}
			\centering
			\includegraphics[width=0.6\textwidth]{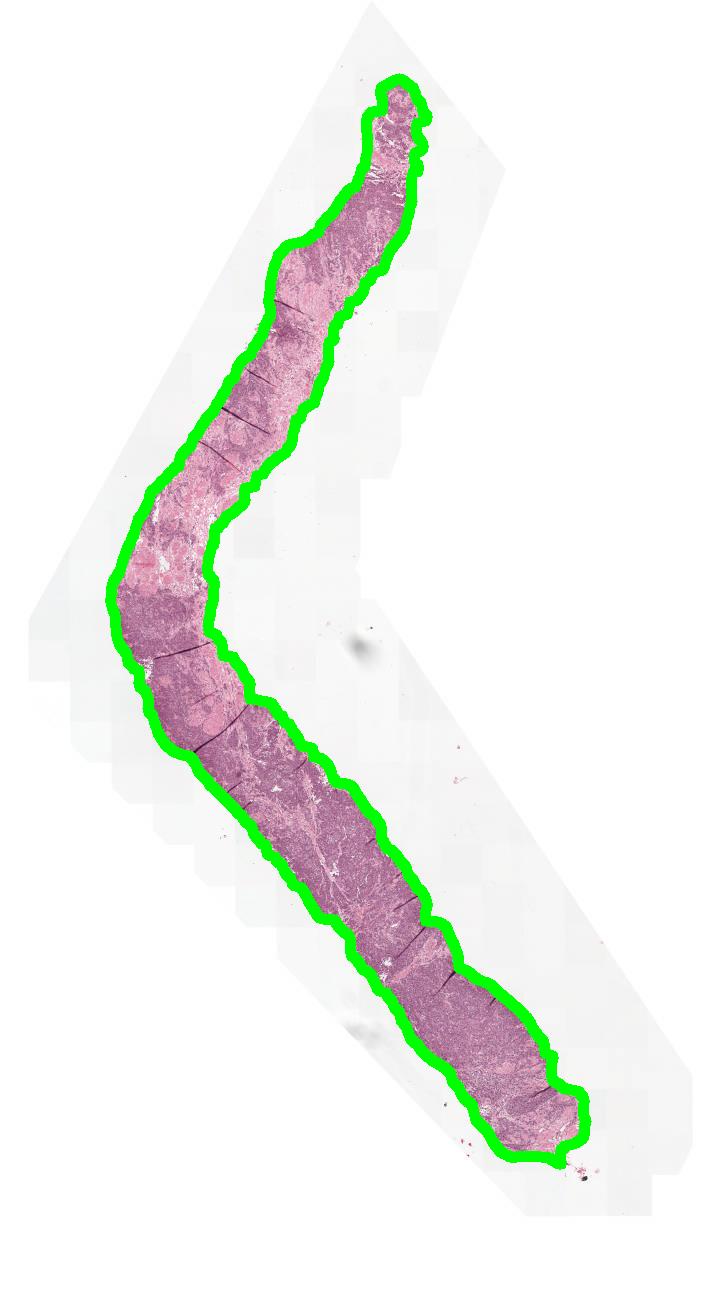}	
			\caption{tissue segmentation}
		\end{subfigure}
		\begin{subfigure}{0.45\textwidth}
			\centering
			\includegraphics[width=0.6\textwidth]{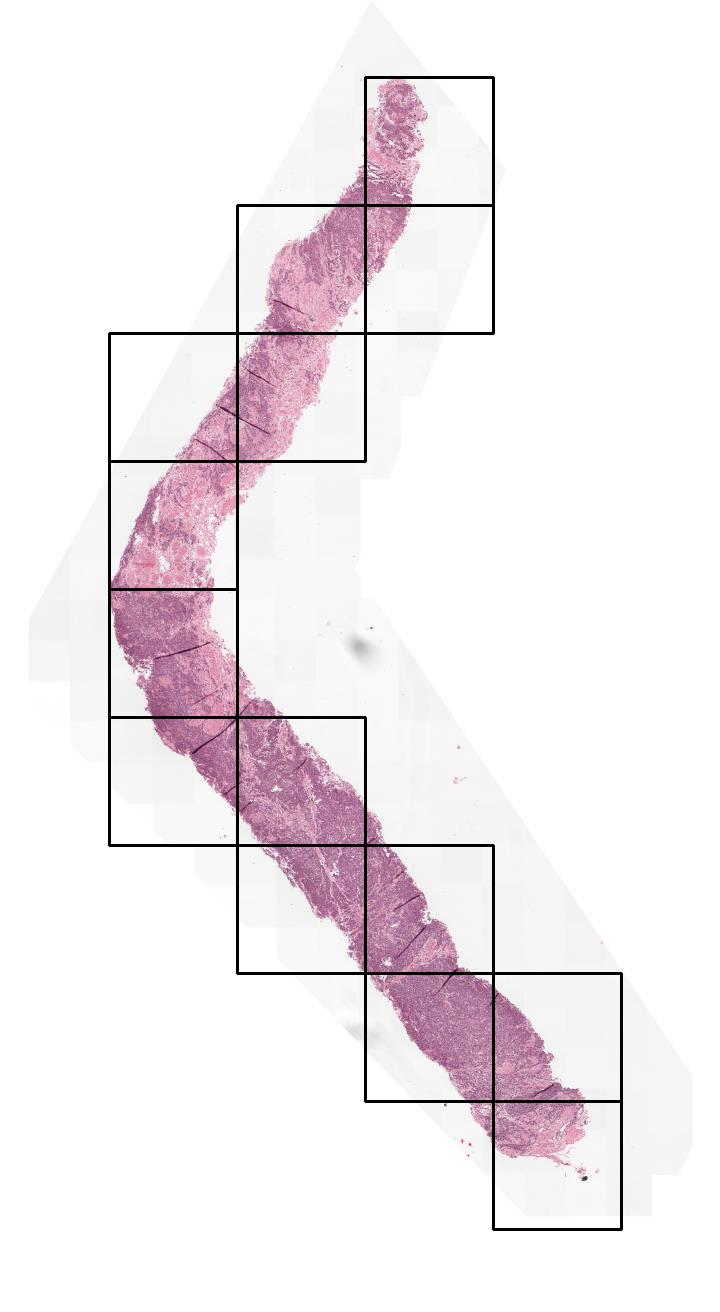}
			\caption{($2048, 2048$) regions extracted at $0.50$ $\mu$m}
		\end{subfigure}
		\caption{Example result of data preprocessing pipeline}
		\label{fig:hs2p}
	\end{figure}

	Figure \ref{fig:hs2p} shows an example result of our tissue segmentation and region extraction algorithm. Regions containing fewer than 10\% tissue were discarded.
	
	\section{Winning Solutions to the PANDA Challenge}
	\label{appendix:winning_panda}
	
	Top-performing methods demonstrated remarkable accuracy. On the public test set, the best submissions achieved quadratic weighted kappa scores ranging from $0.89$ to $0.92$. Similarly, on the private test set, the leading entries scored between $0.92$ and $0.94$. Table \ref{tab:selected_teams} shows the results of the $14$\footnote{there were $15$ teams invited in total, but the scores for the team "UCLA Computational Diagnostics Lab" couldn't be retrieved} teams invited to join the PANDA consortium, as well as an additional entry for our best performing model (obtained by ensembling the predictions of each fold).
	
	\begin{table}[h]
		\caption{Comparison with PANDA consortium teams}
		\label{tab:selected_teams}
		\centering
		\begin{tabular}{cccc}
			\\ [-1.1mm] 
			\toprule
			Team Name & Public Test Score & Private Test Score & Combined Test Score\\
			\midrule
			\texttt{Save The Prostate} & $0.9209$ & $0.9377$ & $\bm{0.9280}$ \\
			\texttt{NS Pathology} & $0.9180$ & $0.9340$ & $0.9272$ \\
			\texttt{PND} & $0.9108$ & $\bm{0.9408}$ & $0.9252$ \\
			\texttt{iafoss} & $0.9179$ & $0.9301$ & $0.9250$ \\
			\texttt{Aksell} & $0.9212$ & $0.9274$ & $0.9247$ \\
			\texttt{vanda} & $\bm{0.9219}$ & $0.9303$ & $0.9215$ \\
			\texttt{ChienYiChi} & $0.9086$ & $0.9324$ & $0.9214$ \\
			\texttt{BarelyBears} & $0.9118$ & $0.9326$ & $0.9204$ \\
			\texttt{rähmä.ai} & $0.9096 $&$0.9262$ & $0.9182$ \\
			\texttt{ctrasd123} & $0.8948$ & $0.9324$ & $0.9165$ \\
			\texttt{Kiminya} & $0.9007$ & $0.9328$ & $0.9164$ \\
			\texttt{Manuel Campos} & $0.8935$ & $0.9296$  & $0.9142$ \\
			\texttt{Dmitry A. Grechka} &	$0.8861$ & $0.9283$ & $0.9105$ \\
			\texttt{KovaLOVE v2} & $0.8889$ & $0.9277$ & $0.9099$ \\
			\midrule
			Local H-ViT & $0.9149$ & $0.9170$ & $0.9161$ \\
			\bottomrule
		\end{tabular}
	\end{table}
	
	In our comparative analysis, a two-sided permutation test on the combined test set between our model and the best performing team (\texttt{Save The Prostate}) shows the difference in performance is not statistically significant ($p = 0.1235$).
	
	\section{Factorized Attention Heatmaps}
	\label{appendix:hm}
	
\begin{figure}[h]
		\centering
		\begin{subfigure}[h]{0.3\textwidth}
				\centering
				\includegraphics[height=1.6\textwidth]{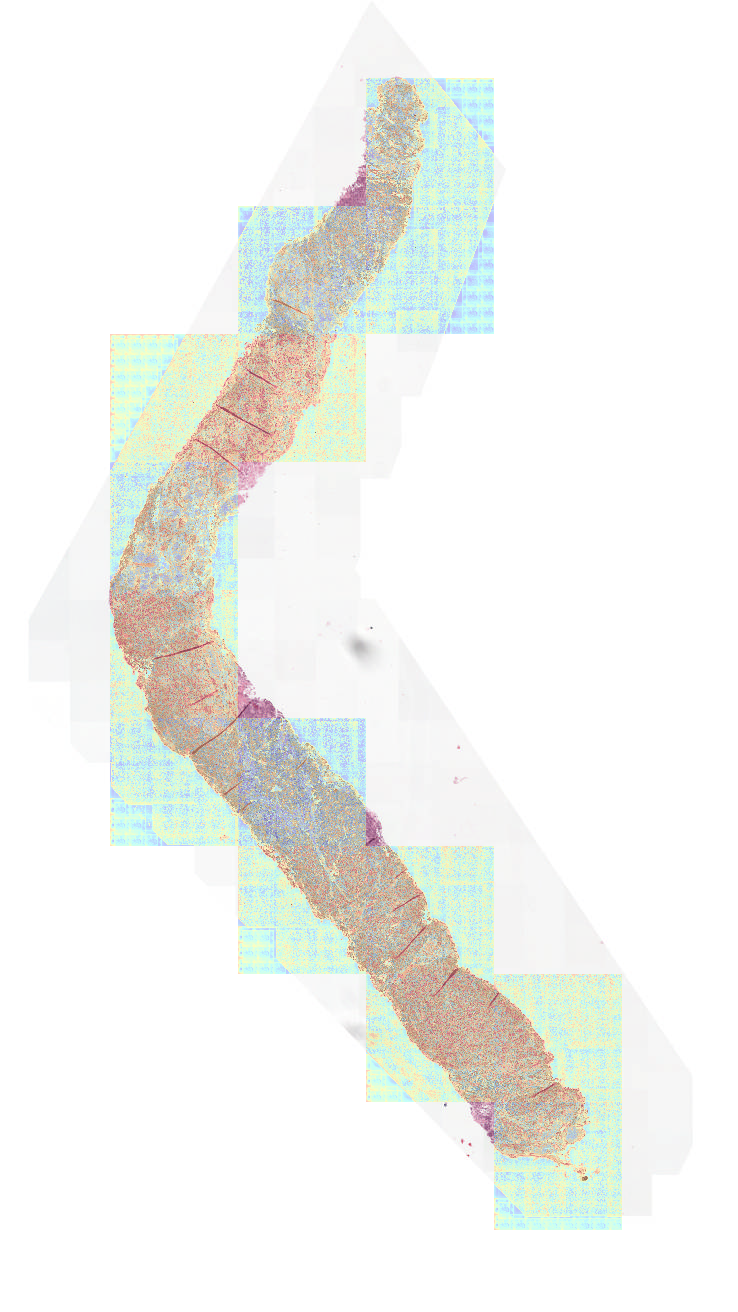}
				\caption{$\gamma = 0.3$}
			\end{subfigure}
		\begin{subfigure}[h]{0.3\textwidth}
				\centering
				\includegraphics[height=1.6\textwidth]{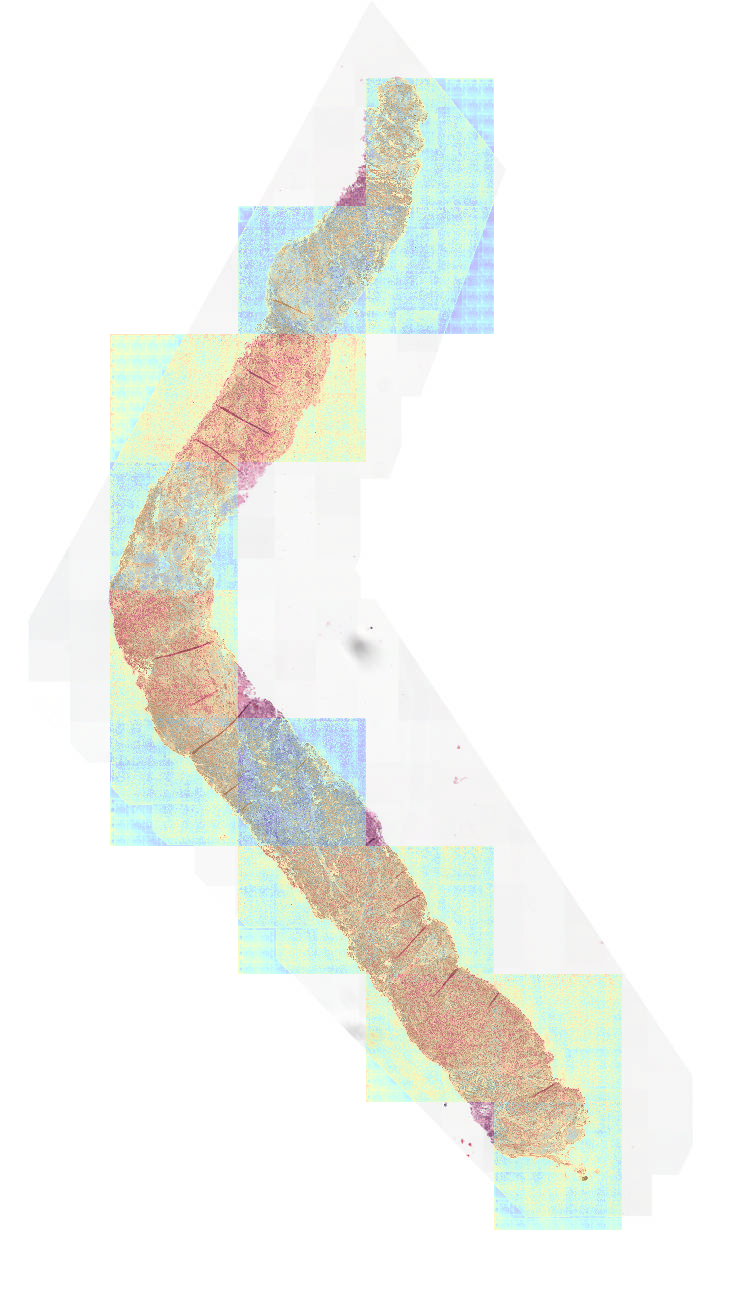}
				\caption{$\gamma = 0.5$}
			\end{subfigure}
		\begin{subfigure}[h]{0.3\textwidth}
				\centering
				\includegraphics[height=1.6\textwidth]{fig/local_hm/r2_p2_gamma=0.7_2048.png}
				\caption{$\gamma = 0.7$}
			\end{subfigure}
		\caption{Local H-ViT factorized attention heatmaps for varying values of parameter $\gamma$}
		\label{fig:hm_gamma}
	\end{figure}

	 Figure \ref{fig:hm_gamma} shows an example of factorized heatmaps for varying values of parameter $\gamma$. The choice of $\gamma$ indirectly controls the granularity of the factorized attention heatmaps: values closer to $0$ produce finer-grained heatmaps focused on patch-level attention, whereas values closer to $1$ yield coarse-grained heatmaps focused on region-level and slide-level attention. Further analysis with expert pathologists is needed to better understand the morphological patterns identified by the model. This collaboration will also help determine the most suitable value for $\gamma$.
	
\end{document}